\documentclass[review]{elsarticle}
\makeatletter
\def\ps@pprintTitle{%
  \let\@oddhead\@empty
  \let\@evenhead\@empty
  \let\@oddfoot\@empty
  \let\@evenfoot\@oddfoot
}
\makeatother
\usepackage{framed,multirow}
\usepackage{natbib}
\usepackage{float}
\usepackage{geometry}
\usepackage{float}
\usepackage{amssymb}
\usepackage{latexsym}

\usepackage{url}
\usepackage{xcolor}
\usepackage{hyperref}
\usepackage{makecell}
\usepackage{lineno,hyperref,color}
\usepackage[figuresright]{rotating}
\usepackage{makecell}
\usepackage[ruled,vlined,linesnumbered]{algorithm2e}

\usepackage{multicol}
\usepackage{listings}

\newcommand\YAMLcolonstyle{\color{red}\mdseries}
\newcommand\YAMLkeystyle{\color{black}\bfseries}
\newcommand\YAMLvaluestyle{\color{blue}\mdseries}

\makeatletter

\newcommand\language@yaml{yaml}

\expandafter\expandafter\expandafter\lstdefinelanguage
\expandafter{\language@yaml}
{
  keywords={true,false,null,y,n},
  keywordstyle=\color{darkgray}\bfseries,
  basicstyle=\YAMLkeystyle,                                 
  sensitive=false,
  comment=[l]{\#},
  morecomment=[s]{/*}{*/},
  commentstyle=\color{purple}\ttfamily,
  stringstyle=\YAMLvaluestyle\ttfamily,
  moredelim=[l][\color{orange}]{\&},
  moredelim=[l][\color{magenta}]{*},
  moredelim=**[il][\YAMLcolonstyle{:}\YAMLvaluestyle]{:},   
  morestring=[b]',
  morestring=[b]",
  literate =    {---}{{\ProcessThreeDashes}}3
                {>}{{\textcolor{red}\textgreater}}1     
                {|}{{\textcolor{red}\textbar}}1 
                {\ -\ }{{\mdseries\ -\ }}3,
}

\lst@AddToHook{EveryLine}{\ifx\lst@language\language@yaml\YAMLkeystyle\fi}
\makeatother

\newcommand\ProcessThreeDashes{\llap{\color{cyan}\mdseries-{-}-}}

\lstdefinestyle{mystyle}{
    basicstyle=\ttfamily\footnotesize,
    breakatwhitespace=false,         
    breaklines=true,                 
    captionpos=b,                    
    keepspaces=true,                 
    numbers=left,                    
    numbersep=5pt,                  
    showspaces=false,                
    showstringspaces=false,
    showtabs=false,                  
    tabsize=2
}
\usepackage{caption}

\usepackage{mdframed}
\usepackage{setspace}
\usepackage{graphicx}
\usepackage{subfig}

\usepackage{wrapfig}

\usepackage{amsmath}

\usepackage{todonotes}

\definecolor{newcolor}{rgb}{.8,.349,.1}

\usepackage{lineno,hyperref}
\modulolinenumbers[5]

\modulolinenumbers[5]

\journal{}

\newcommand{\pyobject}[1]{\color{black}{\texttt{#1}}}

\begin{document}
\date{}
\newcommand\KH[1]{\textcolor{black}{#1}}
\newcommand\KHC[1]{\textcolor{blue}{#1}}

\begin{frontmatter}

\title{Modular Deep Active Learning Framework for Image Annotation: A Technical Report for the Ophthalmo-AI Project}%

\author[add1]{Md Abdul Kadir}
\author[add1]{Hasan Md Tusfiqur Alam}
\author[add1]{Pascale Maul}
\author[add1]{Hans-J\"urgen Profitlich}
\author[add1]{Moritz Wolf}
\author[add1,add2]{Daniel Sonntag}

\address[add1]{German Research Center for Artificial Intelligence (DFKI), Saarbrücken, Germany}
\address[add2]{Oldenburg University, Oldenburg, Germany}

\begin{abstract}
Image annotation is one of the most essential tasks for guaranteeing proper treatment for patients and tracking progress over the course of therapy in the field of medical imaging and disease diagnosis. However, manually annotating a lot of 2D and 3D imaging data can be extremely tedious. Deep Learning (DL) based segmentation algorithms have completely transformed this process and made it possible to automate image segmentation. By accurately segmenting medical images, these algorithms can greatly minimize the time and effort necessary for manual annotation. Additionally, by incorporating Active Learning (AL) methods, these segmentation algorithms can perform far more effectively with a smaller amount of ground truth data. We introduce MedDeepCyleAL, an end-to-end framework implementing the complete AL cycle. It provides researchers with the flexibility to choose the type of deep learning model they wish to employ and includes an annotation tool that supports the classification and segmentation of medical images. The user-friendly interface allows for easy alteration of the AL and DL model settings through a configuration file, requiring no prior programming experience. While MedDeepCyleAL can be applied to any kind of image data, we have specifically applied it to ophthalmology data in this project. 
\end{abstract}
\end{frontmatter}


\section{Introduction}

Optical Coherence Tomography (OCT) has been widely used in biomedical imaging. Ophthalmologists use the segmentation of ocular OCT images for the diagnosis and treatment of eye diseases such as age-related macular degeneration (AMD), diabetic retinopathy (DR), and diabetic macular edema (DME) \cite{dme}. 
In recent years, Deep Learning (DL) based methods have achieved considerable success in medical image segmentation \cite{DL-MI}.
However, their progress has often been constrained as they require large data sets.
Labeling and segmenting medical image data is a time-consuming and expensive process as  domain experts are required to annotate them. 
Active learning can be beneficial for medical image segmentation since it reduces the burden of extensive annotation effort \cite{AL-MI}.

\begin{figure}[!h]
\begin{center}
  \tikzset{every picture/.style={line width=0.75pt}} 

\begin{tikzpicture}[x=0.75pt,y=0.75pt,yscale=-.8,xscale=0.7]

\draw  [fill={rgb, 255:red, 3; green, 49; blue, 104 }  ,fill opacity=1 ] (218.14,294.75) -- (267.26,197.5) -- (431.01,197.5) -- (480.13,294.75) -- (431.01,392) -- (267.26,392) -- cycle ;
\draw   (5,197.5) -- (54.12,100.25) -- (217.87,100.25) -- (266.99,197.5) -- (217.87,294.75) -- (54.12,294.75) -- cycle ;
\draw   (5,392) -- (54.12,294.75) -- (217.87,294.75) -- (266.99,392) -- (217.87,489.25) -- (54.12,489.25) -- cycle ;
\draw   (431.01,197.5) -- (480.13,100.25) -- (643.88,100.25) -- (693,197.5) -- (643.88,294.75) -- (480.13,294.75) -- cycle ;
\draw   (431.01,392) -- (480.13,294.75) -- (643.88,294.75) -- (693,392) -- (643.88,489.25) -- (480.13,489.25) -- cycle ;
\draw   (218.14,100.25) -- (267.26,3) -- (431.01,3) -- (480.13,100.25) -- (431.01,197.5) -- (267.26,197.5) -- cycle ;
\draw   (218.14,489.25) -- (267.26,392) -- (431.01,392) -- (480.13,489.25) -- (431.01,586.5) -- (267.26,586.5) -- cycle ;
\draw  [fill={rgb, 255:red, 3; green, 49; blue, 104 }  ,fill opacity=1 ] (255.45,132.08) -- (251.23,164.13) -- (231.53,127.08) -- cycle ;
\draw  [fill={rgb, 255:red, 3; green, 49; blue, 104 }  ,fill opacity=1 ] (134.72,270.7) -- (151.3,294.43) -- (117.42,293.48) -- cycle ;
\draw  [fill={rgb, 255:red, 3; green, 49; blue, 104 }  ,fill opacity=1 ] (229.39,422.47) -- (253.23,416.68) -- (234.21,454.38) -- cycle ;
\draw  [fill={rgb, 255:red, 3; green, 49; blue, 104 }  ,fill opacity=1 ] (439.11,443.72) -- (442,414.5) -- (460.33,451.18) -- cycle ;
\draw  [fill={rgb, 255:red, 3; green, 49; blue, 104 }  ,fill opacity=1 ] (564.03,319.32) -- (547.42,295.2) -- (580.72,295.32) -- cycle ;
\draw  [fill={rgb, 255:red, 3; green, 49; blue, 104 }  ,fill opacity=1 ] (471.85,159.33) -- (447.72,164.57) -- (466.22,127.36) -- cycle ;

\draw (315.91,278.75) node [anchor=north west][inner sep=0.75pt]   [align=left] {{\textcolor[rgb]{1,1,1}{\textbf{Al Cycle}}}};
\draw (210.18,22.03) node [anchor=north west][inner sep=0.75pt]   [align=left] {\begin{minipage}[lt]{141.93pt}\setlength\topsep{0pt}
\begin{center}
\textbf{1.}\\\textbf{Annotation Tool}\\samples are annotated\\HTTP Request to Server\\{\small update\_annotations\\(annotations)}
\end{center}

\end{minipage}};
\draw (27.18,143.03) node [anchor=north west][inner sep=0.75pt]   [align=left] {\begin{minipage}[lt]{113.33pt}\setlength\topsep{0pt}
\begin{center}
\textbf{6.}\\\textbf{Annotation tool}\\Call preparation tool\\{\small download\_new\_samples()}\\Start annotations
\end{center}

\end{minipage}};
\draw (13,319) node [anchor=north west][inner sep=0.75pt]   [align=left] {\begin{minipage}[lt]{135.01pt}\setlength\topsep{0pt}
\begin{center}
\textbf{5.}\\\textbf{Server}\\HTTP Request to \\annotation tool\\{\small new\_samples(queried\_sample)}
\end{center}

\end{minipage}};
\draw (280,423) node [anchor=north west][inner sep=0.75pt]   [align=left] {\begin{minipage}[lt]{61.49pt}\setlength\topsep{0pt}
\begin{center}
\textbf{4. }\\\textbf{AL Backend}\\ train\\test\\{\small predict()}\\{\small get\_query()}
\end{center}

\end{minipage}};
\draw (476,348) node [anchor=north west][inner sep=0.75pt]   [align=left] {\begin{minipage}[lt]{92.15pt}\setlength\topsep{0pt}
\begin{center}
\textbf{3.}\\\textbf{Server}\\Call AL Backend\\{\small trigger\_AL\_iteration()}
\end{center}

\end{minipage}};
\draw (473,154) node [anchor=north west][inner sep=0.75pt]   [align=left] {\begin{minipage}[lt]{92.7pt}\setlength\topsep{0pt}
\begin{center}
\textbf{2.}\\\textbf{Server}\\Call Datamanager\\{\small update\_annotations()}
\end{center}

\end{minipage}};

\end{tikzpicture}  
\end{center}

\caption{
Design of the AL cycle and the functionality of the individual components:
The first step consists of the annotation of an initial data set through the Annotation Tool (AT).
The annotation information is then forwarded to a Controller, which initiates the update of the data set (2.) and triggers the next AL learning cycle (3.).
The back-end then trains a deep learning model and performs the querying of the non-annotated data (4.).
The top $n$ samples resulting from the querying are then sent to the controller, which in turn sends the information to the AT (5).
With the help of a data preparation tool, the AT then downloads new samples (6.) and generates preliminary annotations with a pre-trained model. 
Those preliminary annotations serve to facilitate the annotation process by proposing labels/segmentation, thus reducing annotation costs.
}

\label{fig:cycle}
\end{figure}
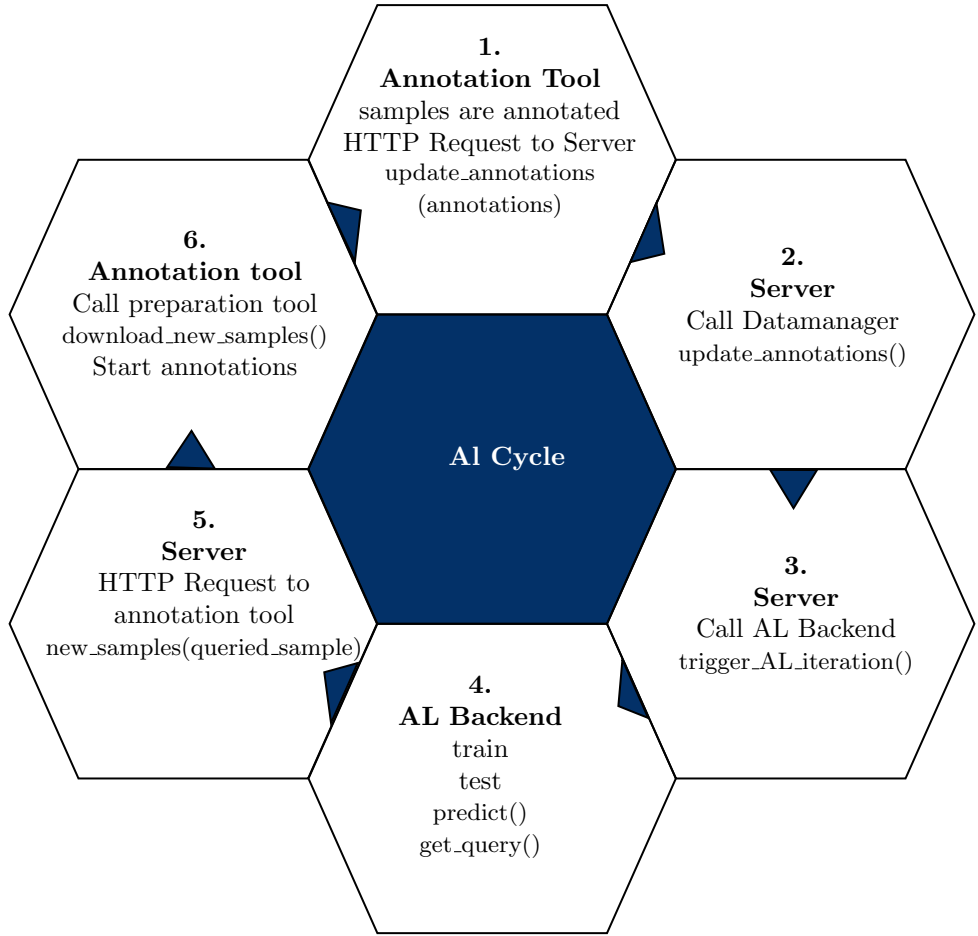

Active Learning (AL) is a paradigm in supervised Machine Learning (ML) where the model interacts with a user to label new data points. 
It is often used in scenarios with a large pool of unlabeled data where the labeling process is expensive. 
By using the ML model to select which examples to learn from, the algorithm can learn a concept with fewer examples than traditional supervised learning \cite{settles2009active}. 
This work focuses on pool-based AL, i.e. uses a fixed data pool to select samples for annotation.
Active learning is an iterative process (Figure \ref{fig:cycle}) that starts with a small, annotated data set to train the ML model. 
This model is then used to query the remaining data from the unlabeled pool. 
Querying refers to the process of assigning scores to individual data points, according to how informative they are. 
The most informative data points are then selected for the next annotation round and added to the training data set. 
The cycle is repeated until the model reaches a given target performance.



The objectives of this work are to create an end-to-end modular AL system for deep learning models, that supports the annotation process, data handling, and the AL iterations. 
It has been developed in the context of the \textbf{Ophthalmo-AI} project (BMBF, see \url{https://www.dfki.de/en/web/research/projects-and-publications/project/ophthalmo-ai)}, which focuses on \textbf{Intelligent, Cooperative Medical Decision Support in Ophthalmology}\footnote{\url{https://www.dfki.de/en/web/news/ohthalmo-ai-started}}.  
To that end, the authors trained a model for the segmentation of retinal OCT images. 
We employed an uncertainty score for the querying, since the available unannotated data contains many healthy eyes and a large set of medical markers for the diseases.
Previous projects in this area do not include an end-to-end solution, do not focus on deep learning, and have to be integrated programmatically.
Moreover they focus on classification, while our framework offers support for segmentation.

\subsection{The Ophthalmo-AI Project}
The developments in the Ophthalmo-AI project are intended to support ophthalmologists with an intelligent assistance system that can help them make the best possible therapy decision by making a correct diagnosis based on both image and clinical data. 
To generate comprehensible suggestions for medical staff, the AI system will first label biological structures and pathological features in the image data. Then special AI models derive diagnoses from the image findings and other information from the patient file, make therapy suggestions and predict the success of the therapy. Interactive machine-learning methods are used to integrate the doctors' knowledge of the case in question into the process.  A large volume of treatment data collected and processed on the special data integration platform "XplOit" \cite{xploit} is used for system development. The resulting augmented intelligence system will be tested in clinical demonstrators for its practical suitability for macular degeneration and diabetic retinopathy. 

The project focuses on the two most common causes of blindness in people over fifty: age-related macular degeneration and diabetic retinopathy. AMD is caused by aging processes in the central retinal cells (macula = point of sharpest vision). 
Everyone has a buildup of waste products in this area throughout life, and some have irregular blood vessels growing from the choroid under the retina. They leak fluid and blood, causing symptoms of "distortions" and progressive vision loss ("wet" form of "late" AMD). Diabetic retinopathy can occur as a result of diabetes. Here, too, blood components can escape into the retinal tissue. From this, pathological new blood vessels – so-called “proliferations” – can grow on the retina. Heavy bleeding into the eye can occur from these very vulnerable vessels, which can cause retinal detachment. Retinopathy often leads to DME, a swelling in the area of the macula, which can lead to visual disturbances.
The layered representation of the retina, the "optical coherence tomography" (OCT), makes it possible to detect the finest changes and swellings, such as the presence of irregular blood vessels in the area of the macula. This is important for the diagnosis and therapy monitoring of both diseases. New drugs are now available that suppress the growth of abnormal blood vessels and reduce swelling in the macula. These drugs are injected directly into the vitreous cavity of the eye (intravitreal drug delivery, IVOM). The IVOM must be repeated at different intervals, usually monthly, until the vision has stabilized.

In the first project phase, both textual annotations in the patient files and annotations of the retinal structures (especially layers and drusen) are made using specially developed tools. To optimize the required working time, a tool already supports the doctors with segmentation models and special graphic algorithms. To further reduce the time-consuming annotation process, AL is used, which specifically selects those images that promise the greatest benefit in the continuous improvement of the models. In the second step, the resulting neural networks serve to support diagnoses and therapy suggestions and predict the success of the therapy. In order to enable the doctors to understand the system's decisions, in addition to the assessment of an existing disease, explanations of the results are also given in the form of interactive segmentation maps of the characteristics relevant to the diagnosis.


\section{Related Work}

Active learning is a strategy that focuses on the annotation of highly informative samples to improve model performance in a cost-effective manner. This selection is based on factors such as uncertainty \cite{lee2018robust}, data distribution \cite{samrath2019variational}, expected model change \cite{dai2020suggestive}, among others \cite{bai2022discrepancy}. Uncertainty can be measured through probabilistic analysis of predictions by selecting instances with the lowest confidence \cite{lee2018robust,joshi2009multi}, or by calculating class entropy \cite{luo2013latent}.

Applications of uncertainty-based approaches in deep neural networks have been explored in a few studies \cite{siddiqui2020viewal}. Notably, \citet{gal2017deep} proposed an uncertainty estimation method using dropout-based Monte Carlo (MC) sampling. This approach involves multiple forward passes with dropout throughout various layers, generating uncertainty during inference. Ensemble-based methods, which measure uncertainty based on the variance between predictions from numerous models, have also been utilized \cite{nath2020diminishing,yang2017suggestive,sener2017active}.

Active learning methods have been utilized for a range of segmentation tasks. \citet{gorriz2017cost} extended the Cost-Effective Active Learning (CEAL) algorithm \cite{wang2016cost} for Melanoma segmentation, wherein they chose both high and low-confidence samples for annotation. \citet{mackowiak2018cereals} employed a region-based selection approach with MC dropout to estimate model uncertainty, thereby reducing annotation costs. Moreover, \citet{nath2020diminishing} proposed an ensemble-based approach, using multiple active learning frameworks that are jointly optimized. They adopted a query-by-committee strategy for sample selection. \citet{siddiqui2020viewal} also suggested an active learning framework for multi-view dataset segmentation, quantifying model uncertainty based on the Kullback-Leibler divergence of posterior probability distributions.

\citet{gaillochet2023active} proposed a representative learning-based active learning strategy for MRI prostate segmentation, where samples are selected for annotation at the batch level during training, rather than using typical stochastic batches. This innovative approach contributes to the speed of training convergence and increases model robustness by concentrating on specific data points.
Despite the progress in Active Learning for various applications, there remains a scarcity of comprehensive studies dedicated to Active Learning for OCT segmentation, with \cite{li2021unsupervised} being one of the few one can mention. This work by \citet{li2021unsupervised} necessitates the use of foundation models, as per \citet{khan2020survey}, requiring pre-training on large-scale datasets from similar domains. However, the practical application of this methodology could be challenged due to the difficulty in data collection, particularly considering the strict constraints tied to data privacy. \citet{kadir2023edgeal}  propose EdgeAL technique, which uses edge information from OCT images as an a priori to improve the quality of queries in Active Learning for Deep Segmentation model training. The main concept is to take advantage of predictions' uncertainty as they cross semantic region borders in the input images.

The active learning framework proposed in this paper is designed to accommodate and integrate all previously mentioned active learning algorithms effectively. This encompassing approach facilitates seamless integration with a wide range of deep learning architectures and configurations, contributing to its versatility and broad applicability in the field.

\section{OCT: Optical Coherence Tomography}


OCT (optical coherence tomography) is a technique used to get a layered representation of the retina. 
This enables sectional images of the back of the eye to be taken in high resolution by using a laser light. The retina reflects this weak laser light differently; these reflections are measured and converted into detailed images by a computer system. These images provide a precise insight into the finest structures and changes in the retinal layers.

\begin{figure}%
\centering
\subfloat[][]{\includegraphics[trim={0 1mm 0 3mm},clip, width=0.47\textwidth]{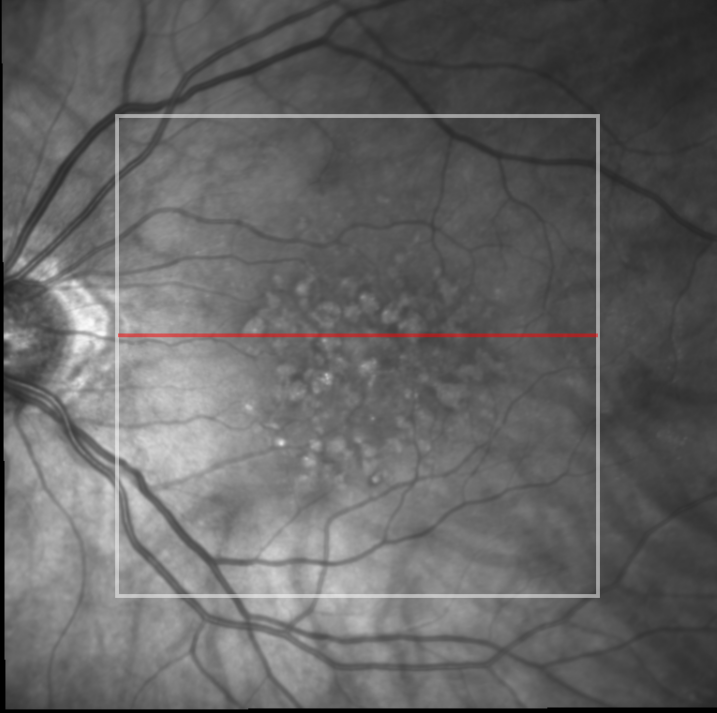}}%
\qquad
\subfloat[][]{\includegraphics[width=0.47\textwidth]{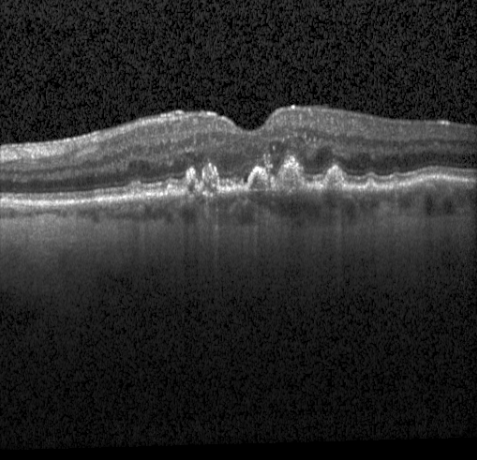}}
\caption{OCT: a) SLO with the positioning of the exposure on the retina (white square) and the position of a slice (red line), and b) corresponding slice}%
\label{fig:oct}%
\end{figure}

Besides a top-level view (called SLO) a number of slices (called b-scans) are recorded that allow a view under the surface of the retina. Each of these slices shows the individual layers of the retina, whose changes are essential for the assessment of the diseases under consideration and the choice of the right therapy (Figure \ref{fig:oct}).
The implemented tool supports the annotation of up to 20 features within each slice, including the different layers, fluid collections and other bio-markers (for more details see section \ref{sec:iuiannotool}).
The SLO image is saved in a file format called VOL together with the slices and metadata about the procedure, including a patient identifier, the number of slices, distance to the eye, software or hardware versions, etc.

\section{Project Specific Objectives}

The objective of this project phase is to create an AL end-to-end system that can easily be configured and extended by AL users of all experience levels. 
To that end we build our architecture on top of frameworks that are used by expert machine learning practitioners (PyTorch), but can easily be configured by non-experts.
The tool 
does not only handle training but is also connected to the data annotation tool \cite{wolf-2020-human} and handles the whole process from data set compilation with the help of AL, data annotation up to model training.
The benefits for the users are:
No manual integration of system components from very different backgrounds (e.g., data annotation vs. model training) is needed.
Moreover, the user has an easily scalable system at hand that can be used for online and offline AL experiments.
Since this framework was built as a decentralized system, individual components can be replaced if necessary.

\section{Architecture}

\begin{figure}[!h]
\begin{center}
  \includegraphics[width=\textwidth]{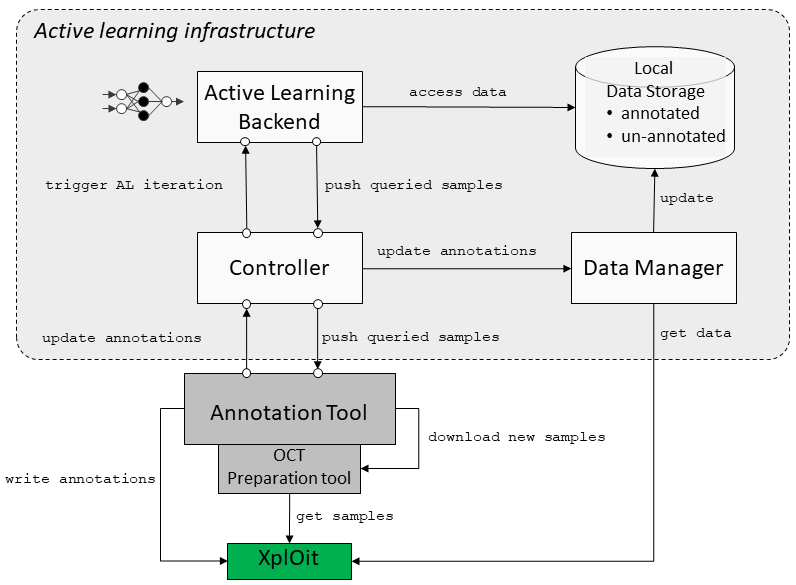}
\caption{Architecture of the Active Learning Infrastructure}  
\label{block}
\end{center}

\end{figure}


MedDeepCyleAL is designed to optimize the process of annotating and training deep learning models. The system's four components (Figure \ref{block}), namely the Annotation Tool (AT), Controller, Data Manager, and Active Learning Backend, work together seamlessly to create a decentralized and powerful platform for enhancing the development of interactive deep learning tools.


The initial dataset is annotated with the help of an expert, without employing any AL strategy. This annotated dataset is then forwarded to the Controller. 
It plays a pivotal role in ensuring the proper execution of the AL cycle by coordinating and initiating the various modules within the MedDeepCycleAL system.

Upon receiving the annotated dataset, the Controller activates the Data Manager, which is responsible for downloading and managing the dataset used for model training, as well as maintaining the pool of unannotated data. This process ensures that the training data stays up-to-date and relevant to the data available in the data warehouse, allowing for more accurate model development.

Once the dataset is prepared, the Active Learning Backend proceeds to train a deep learning model based on the model configuration. The trained model is then employed to rank the remaining unannotated data according to the selected AL query method. The Controller picks the most important sample for further annotation based on the ranking and presents it to the expert annotation tool interface for annotation. After the expert annotates the queried samples, the controller initiates the training cycle again. This strategic approach to data selection enables a more efficient use of resources and improved model performance with a limited amount of annotated data.

\subsection{Controller}
The controller is in charge of coordinating all other components of MedDeepCyleAL.
The controller is a Flask server that, in line with the typical AL process, waits for the annotation tool to deliver some annotations in order to trigger the data manager to create the initial data set. 
With the initial data set, the controller then triggers the \textit{active learning backend} to train a DL model on the initial data set, which is then used to rank the remaining data in the pool of unannotated data points. 
The top $n$ results from that list are then sent to the annotation tool for the next AL iteration. With the resultant annotated data, the \textit{ontroller} forwards to the next active learning cycle.

\subsection{Data-Manager}

The Data Manager is developed as an independent module to support the annotation tool and machine learning backend architecture. It provides APIs to access and process patient's treatment data from the \pyobject{XplOit}\footnote{https://ophthalmo-ai.fraunhofer.de} data server. The \textit{controller} module accesses the data manager, and its functionalities can be divided into three main parts: 1) Securely downloading required patient treatment data from the server to the local development machine, 2) process the raw data, which are stored as a large Volume file, and 3) extract unannotated OCT scan slices and other metadata, including patients' health records (EHRs), diagnosis, and treatments. The Data Manager also contains functionalities to store and access the annotations provided by the \textit{annotation tool} for OCT images. This module also provides APIs to access training/validation/test data sets, which are accessed by \textit{active learning backend} via the controller module. The main APIs of the Data Manager class and their functionalities are briefly described below:

\begin{itemize}
    \item \pyobject{initialize(root\_dir:Path, database:DB)}: Initializes the \pyobject{DataManager} class using the tables in the database containing patient attributes. Required patient files are stored in Volume format (\pyobject{.VOL}) are downloaded using the \pyobject{Xploit} REST API. 
    
    \item \pyobject{initialize\_with\_files(file\_path:Path) -> None}:  The \pyobject{DataManager} class can also be initialized using this function, if the required files are already downloaded from the server and stored in the local system.
    
    \item \pyobject{get\_unannotated\_set() -> List[Dict]} : Returns a list of dictionary items containing the path of unannotated OCT images and associated meta data.
    
    \item \pyobject{remove\_from\_unannotated\_set(file\_list:List[Dict]) -> None}: Removes an object from the unannotated set when a batch of OCT slices are annotated.
    
    \item \pyobject{update\_annotations(annotations:Dict[Dict]) -> None}: Given the annotation values from the Annotation Tool for an OCT image, this function produces the associated semantic masks and stores it in an image file. 
    
    \item \pyobject{get\_annotated\_set() -> List[Dict]}: Returns a list of dictionary items containing the path to annotated OCT images and their corresponding annotated masks. 
    
    \item \pyobject{get\_dataloader(DATA\_ENUM:str) -> list[Dict]}: Returns the pre-processed data set for machine learning model training. Input argument is a literal of three choices \pyobject{DATA\_ENUM=[`train',`validation',`test']}.
\end{itemize}

\subsection{Active-Learning Backend}
We utilized a versatile deep learning backend tool, developed on the PyTorch-based framework VISSL, that can be easily reconfigured and reused for training, testing, validation, and prediction. There are multiple benefits to using this framework, as it enables us to construct a deep learning model, establish hyperparameters, and train the model with a structured and adaptable YAML (Listings \ref{code:model} and \ref{code:data} are two example snippets taken from the YAML file) blueprint file. To enable active learning in our machine learning model, we have integrated the modAL framework, which is specifically designed for active learning. However, since our model also requires segmentation functionality, we have added an additional layer on top of the VISSL and modAL integration. This is necessary because VISSL was originally developed for unsupervised model training, and does not include functionalities for segmentation model training. Additionally, the algorithms built into modAL only support classification tasks, and not all active learning algorithms.

To enable segmentation functionality in our model, we have built additional features on top of the VISSL and modAL integration. This enables our model to perform active learning for segmentation tasks, which was not previously possible with the existing frameworks. By combining the strengths of modAL and VISSL with our new segmentation features, we are able to create a powerful and flexible machine learning model that can be trained using active learning techniques.

\begin{center}
\begin{multicols}{2}
\captionof{lstlisting}{Snippet of the YAML configuration file for the model, loss function and optimizer initialization}
\begin{lstlisting}[numbers=none]
MODEL:
    FEATURE_EVAL_SETTINGS:
        EVAL_MODE_ON: False
    TRUNK:
        NAME: unet
        UNET:
            n_channels: 1
            bilinear: True
    WEIGHTS_INIT:
      PARAMS_FILE: data/unet_best.pth

LOSS:
    name: dice_loss
    dice_loss:
      softmax: True
      ignore_index: -1
OPTIMIZER:
    name: sgd
    momentum: 0.9
\end{lstlisting}

\label{code:model}

\columnbreak
\captionof{lstlisting}{Snippet of the YAML configuration file for data loading, transformation and augmentation}

\begin{lstlisting}[,numbers=none]
DATA:
    NUM_DATALOADER_WORKERS: 1
    TRAIN:
        DATA_SOURCES: [disk_filelist]
        LABEL_SOURCES: [disk_filelist]
        DATASET_NAMES: [seg_data]
        BATCHSIZE_PER_REPLICA: 3
        TRANSFORMS:
        - name: RandomResizedCrop
          size: 224
        - name: ToTensor
        - name: Normalize
          mean: [0.485]
          std: [0.229]
        MMAP_MODE: True
        COPY_TO_LOCAL_DISK: False
        COPY_DESTINATION_DIR: ""
        DATA_LIMIT: 3
        COLLATE_FUNCTION: msk_collator
\end{lstlisting}

\label{code:data}

\end{multicols}
\end{center}

Our YAML configuration file outlines six essential aspects of our active learning tool: ACTIVE LEARNING, MODEL, DATA, METERS, OPTIMIZER, DISTRIBUTED, and MACHINE. 
\begin{enumerate}
    \item ACTIVE LEARNING defines the AL algorithm and hyperparameters to be used for the tool.
    \item  MODEL specifies the deep learning model and its associated hyperparameters.
    \item  DATA outlines the data loading approach and the transformations to be considered during training and testing.
    \item  OPTIMIZER specifies the optimization algorithm and its additional hyperparameters.
    \item  METERS determines the performance metrics to be assessed during evaluation.
    \item  DISTRIBUTED and MACHINE define the hardware configuration for training and testing.
\end{enumerate}

This modular configuration of the YAML file allows us to flexibly add new functionalities to our active learning tool, streamlining the process of integrating updates and improvements.

\subsubsection{Segmentation network}

Deep learning models requires a large set of annotated data for training. Instead of training a segmentation model from the scratch we have adopted transfer learning \cite{weiss2016survey} strategy for our segmentation model. We trained segmentation model using a publicly available OCT dataset of similar task namely \textit{Annotated Retinal OCT Images Database} (AROI) \cite{melinvsvcak2021annotated}. The dataset consist of 1136 annotated OCT scans of 24 patients suffering from age-related macular degeneration disease. Each scans has annotations for three retinal layers and three retinal fluid layers. We train different state-of-the art foundation models for segmentation using AROI dataset including U-Net \cite{ronneberger_2015}, U-Net++ \cite{zhou_2018}, Attention U-Net \cite{oktay2018attention} and Y-Net \cite{dme}. Table \ref{tab:seg_res} compares the segmentation results for different architectures on AROI dataset. 

\begin{figure}
  \centering
  \includegraphics[width=\textwidth]{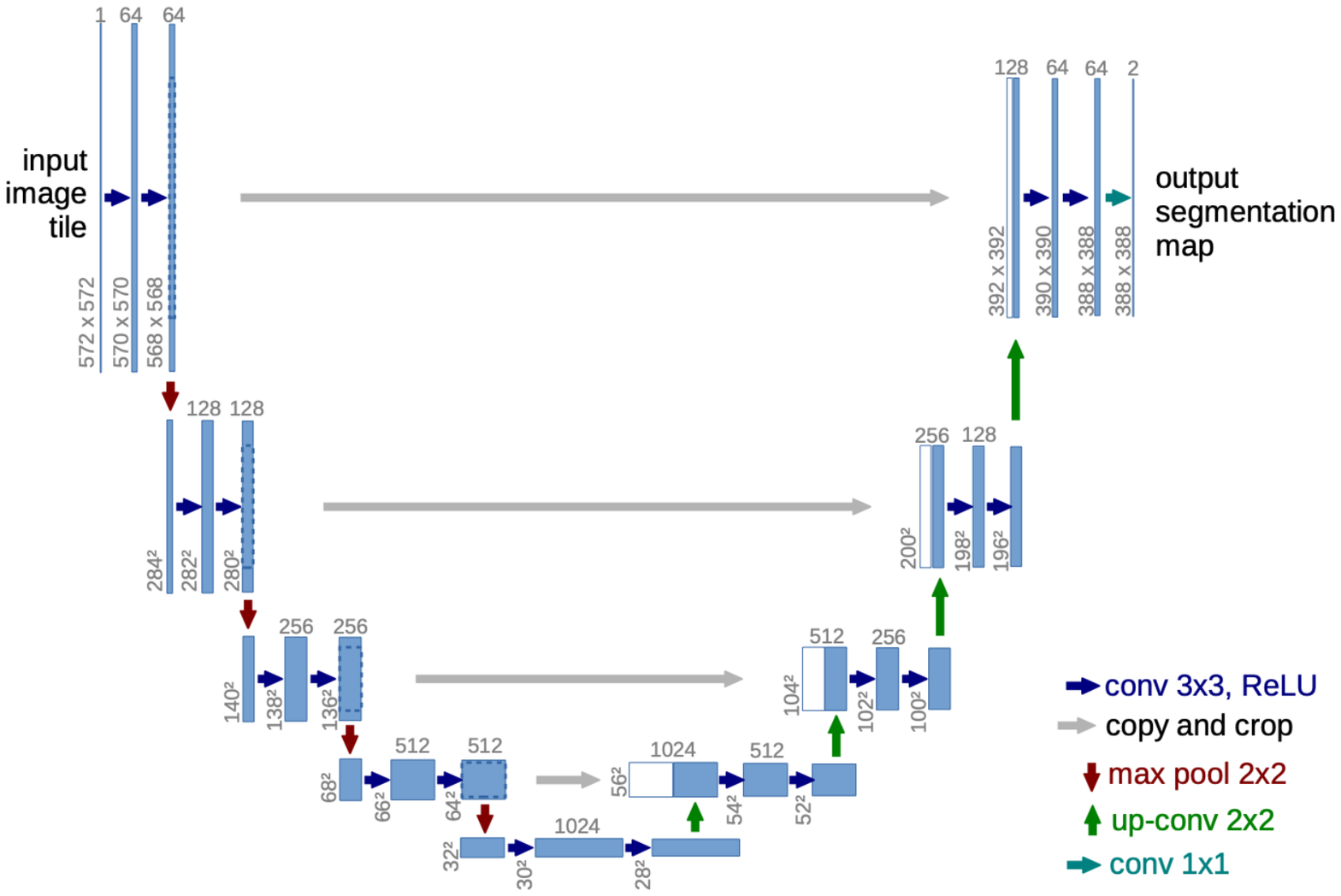}
  \captionsetup{justification=centering}
  \caption{Visual representation of U-net architecture \cite{ronneberger_2015}}
  \label{fig:unet}
\end{figure}

The segmentation network utilized in this tool is based on the U-Net (Figure \ref{fig:unet}) architecture, as described by \citet{ronneberger_2015, zhou_2018}. The network comes with pre-trained weights from \citet{melinscak_2021}. In addition to U-net, we also conducted experiments using Y-net (YN)\cite{farshad2022net} and DeepLab-V3 (DP-V3) \cite{siddiqui2020viewal} with ResNet and MobileNet backbones [10]. Table \ref{tab:5foldcv} presents the performance of various active learning algorithms on 12\% annotated data for different architectures.

\begin{table}[ht!]
\centering
\caption{Class-wise Results for different Segmentation Models on AROI dataset \cite{melinvsvcak2021annotated}}
\scalebox{0.90}{
\begin{tabular}{|l|c|c|c|c|c|c|c|c|}
\hline
            Architecture    & Above ILM & ILM-IPL/INL & IPL/INL-RPE & RPE-BM & Under BM & PED   & SRF   & IRF   \\ \hline
U-Net           & 0.995     & 0.950       & 0.923       & 0.669  & 0.988    & 0.638 & 0.513 & 0.480 \\
Attention U-Net & 0.995     & 0.899       & 0.890       & 0.476  & 0.988    & 0.533 & 0.372 & 0.037 \\
U-Net ++        & 0.992     & 0.944       & 0.924       & 0.641  & 0.986    & 0.622 & 0.487 & 0.419 \\
Y-Net           & 1.000     & 0.970       & 0.950       & 0.640  & 0.990    & 0.630 & 0.520 & 0.980 \\ \hline
\end{tabular}}
\label{tab:seg_res}
\end{table}

\subsubsection{Sampling}
Annotating large amounts of segmentation data can be tedious. To reduce annotation effort, active learning is necessary. Several algorithms exist for active learning selection methods in semantic segmentation, including  softmax margin \textbf{(MAR)} \cite{joshi2009multi}, softmax confidence \textbf{(CONF)} \cite{wang2016cost}, softmax entropy \textbf{(ENT)} \cite{luo2013latent}, MC dropout entropy \textbf{(MCDR)} \cite{gal2017deep}, Core-set selection \textbf{(CORESET)} \cite{sener2017active}, \textbf{(CEAL)} \cite{gorriz2017cost}, and regional MC dropout entropy \textbf{(RMCDR)} \cite{mackowiak2018cereals}, maximum representations \textbf{(MAXRPR)} \cite{yang2017suggestive}, and random selection \textbf{(Random)}.

\begin{table}[ht!]
\centering
\caption{The Table summarizes 5-fold cross-validation results (mean dice) for active learning methods and EdgeAL on the Duke dataset. EdgeAL outperforms other methods, achieving 99\% performance with just 12\% annotated data.}
\label{tab:5foldcv}
\scalebox{0.9}{
\begin{tabular}{|l|c|c|c|c|c|c|c|c|}
\hline
GT(\%) & RMCDR & CEAL &  CORESET & \textbf{EdgeAL} & MAR & MAXRPR  \\ \hline
2\%       & \textbf{0.40 \textpm 0.05} & \textbf{ 0.40 \textpm 0.05} &  0.38  \textpm 0.04  &  \textbf{0.40 \textpm 0.05}  & \textbf{0.40 \textpm 0.09} & \textbf{0.41 \textpm 0.04}   \\
12\%      & 0.44 \textpm 0.04 & 0.54 \textpm 0.04 & 0.44 \textpm 0.05   & \textbf{0.82 \textpm 0.03}  & 0.44 \textpm 0.03 & 0.54  \textpm 0.09  \\
22\%      & 0.63 \textpm 0.05 & 0.54 \textpm 0.04 &  0.62 \textpm 0.04   & \textbf{0.83 \textpm 0.03}  & 0.58 \textpm 0.04  & 0.67 \textpm 0.07    \\
33\%      & 0.58 \textpm 0.07 &0.55 \textpm 0.06 & 0.57 \textpm 0.04   & \textbf{0.81 \textpm 0.04}   & 0.67 \textpm 0.03 & 0.61  \textpm 0.03     \\
43\%      & 0.70 \textpm 0.03 & 0.79 \textpm 0.03 &0.69 \textpm 0.03   & \textbf{0.83 \textpm 0.02}  & 0.70 \textpm 0.04 & 0.80 \textpm 0.04   \\
\hline
100\%      & 0.82 \textpm 0.03 & 0.82 \textpm0.03  &  0.82 \textpm 0.03  & 0.82 \textpm 0.02  & 0.83 \textpm 0.02 & 0.83 \textpm 0.02 \\\hline
\end{tabular}}
\end{table}

We conducted a five-fold cross-validation to evaluate the performance of each active learning algorithm. Our findings suggest that EdgeAL and MAXRPR are the most effective in selecting the best queries for training with a minimal amount of annotation. We presented the results in Table \ref{tab:5foldcv}.


\subsection{Annotation Tool}

After the backend finishes training and querying, the controller sends the queried samples to the annotation tool.
The annotation tool calls the OCT preparation tool which downloads OCTs from XplOit, adding additional random OCTs and loads data into annotation tool. 3 slices per OCT are chosen, the middle and two randomly on either half of the OCT. Doctors are then annotating these slices with the tool. Finished annotations are sent back to the controller.
They are also sent to XplOit for storage and access for use by project partners.

\section{Intelligent User Interfaces} \label{sec:iuiannotool}

\subsection{The Annotation Tool}
\begin{figure}
    \begin{center}
         \includegraphics[width=.8\textwidth]{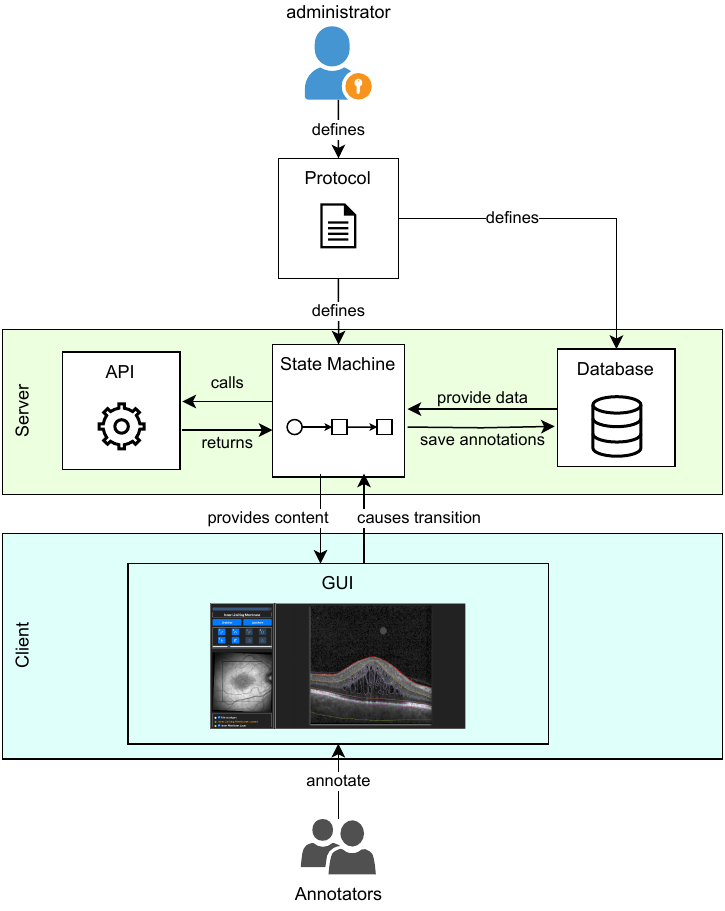}
    \caption{Basic architecture of HUMAN: Administrators define a protocol that is translated into a state machine. The State Machine interacts with API, Database and Client GUI. Annotators annotate using the GUI.}
    \label{fig:human_architecture}
    \end{center}
   
\end{figure}

\begin{figure}[h!]
\centering
\includegraphics[width=\textwidth]{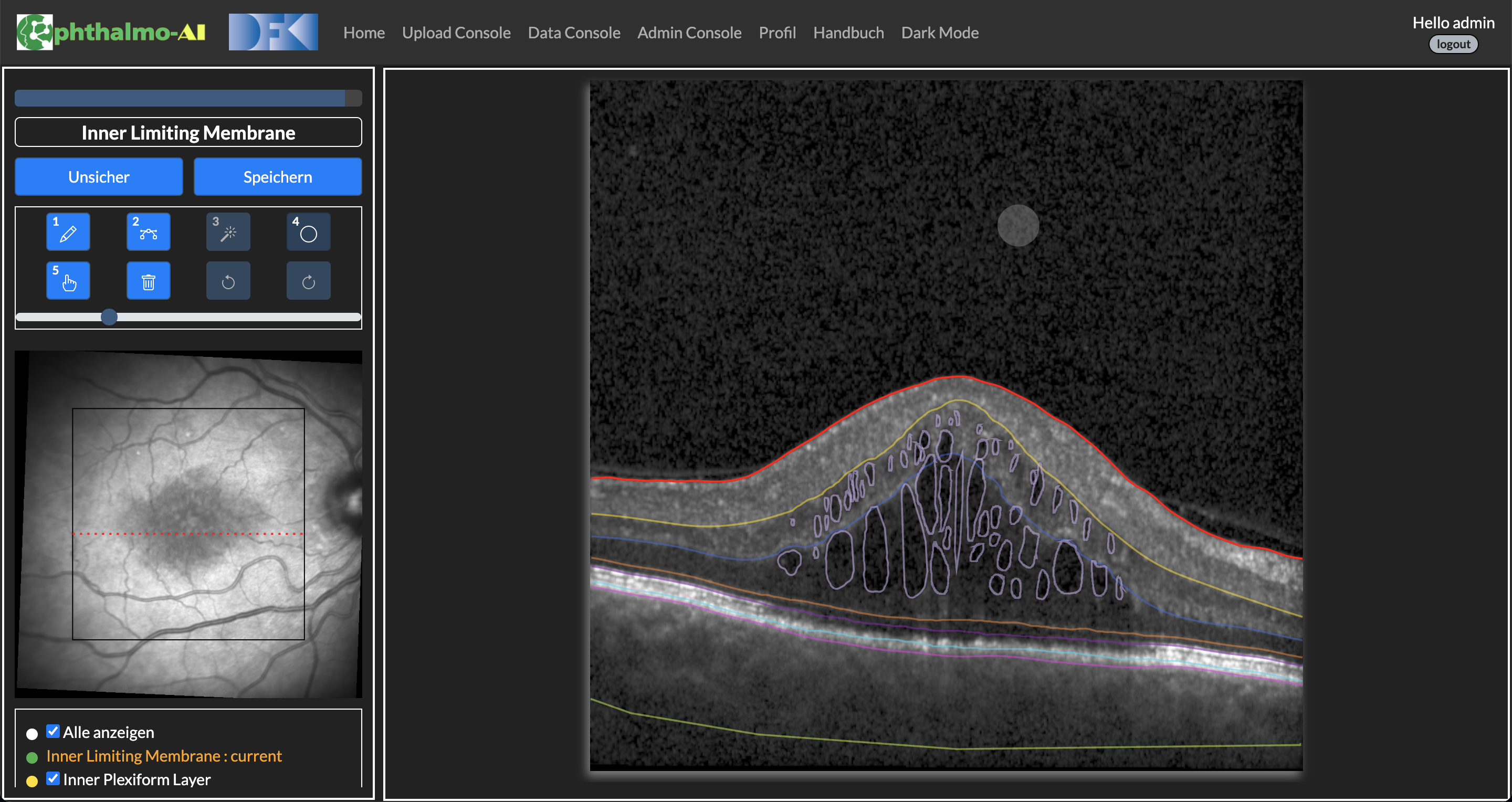}
\caption{Example OCT annotation in HUMAN. Left side, from top to bottom: a progress bar, the current layer or fluid to be annotated, continue and uncertain buttons, drawing tools, the SLO and a list summarizing all annotations. Right side: The slice in which the annotation is to be done. The current annotation is marked red, the others are transparent and colored corresponding to the mark in the summary list.}
\label{fig:human_ui}
\end{figure}

The annotation tool that is used for image annotations is a fork of HUMAN
\cite{wolf-2020-human}. Its main goal was to be as flexible and adaptable to a variety of annotation tasks as possible. The name HUMAN stands for \textbf{H}ierarchical \textbf{U}niversal \textbf{M}odular \textbf{AN}notator, reflecting the main features of this tool: It provides a variety of modular question types like multiple-choice, yes-no or setting bounding boxes that can be arranged via an internal state machine. These modules can be self contained or rely on each other, i.e. there can be a follow up task depending on a previous annotation. E.g. for semantic image labeling, one task could be drawing one or multiple bounding boxes, the next task could be labeling these with a predefined object name. The architecture (Figure \ref{fig:human_architecture}) also allows the input for a task to come from another source, e.g. raw data or a classification model. For the example above a segmentation model could then provide the bounding box coordinates for the bounding box labeling task. The tool is implemented as a native JavaScript web application (Figure \ref{fig:human_ui}) with a Flask server backend. This has the advantage of better compatibility with various operating systems, parallel annotations and automated annotation distribution to different users. The procedures for setting up a data annotation protocol and annotating data instances are described in algorithm \ref{algorithm1}. The administrator must launch a server and create a state machine protocol before the algorithm may begin. In the \ref{annotationtool}, a sample state machine protocol is given. All the annotated information is stored in an SQLite database. \ref{workflow} provides an overview of the tool's workflow along with some example annotations.
\newpage




\begin{algorithm}
\SetAlgoLined
\KwResult{Annotated data instances}
\caption{Protocol Setup and Data Annotation}
\label{algorithm1}

\textbf{Step 1:} Start the server\;
\textbf{Step 2:} Administrator defines a state machine protocol\;
\Indp
\textit{YAML file with task configuration and chaining}\;
\Indm
\textbf{Step 3:} Upload raw data into the database\;
\Indp
\textit{Manually via Upload Console or automatically via OCT Preparation Tool}\;
\Indm
\textbf{Step 4:} Register users for data annotation\;
\While{Not all data instances are annotated}{
    \textbf{Annotation Process:}\;
    \Indp
    User logs in\;
    State machine assigns a data instance\;
    Perform annotations\;
    State machine transitions until end state\;
    Annotations are saved in the database\;
    \Indm
    Assign new data instances for annotation\;
}
\end{algorithm}


\subsubsection*{OCT Specific Changes}

For efficient annotations of OCT images, some changes were needed. For OCT annotations it is often necessary to jump back and forth between tasks. This made it necessary to abandon the hierarchical features of HUMAN in favor of the possibility to jump between states. 


\subsection{Diagnostic Decision Support Prototype}

\begin{figure}
  \centering
  \includegraphics[width=\textwidth]{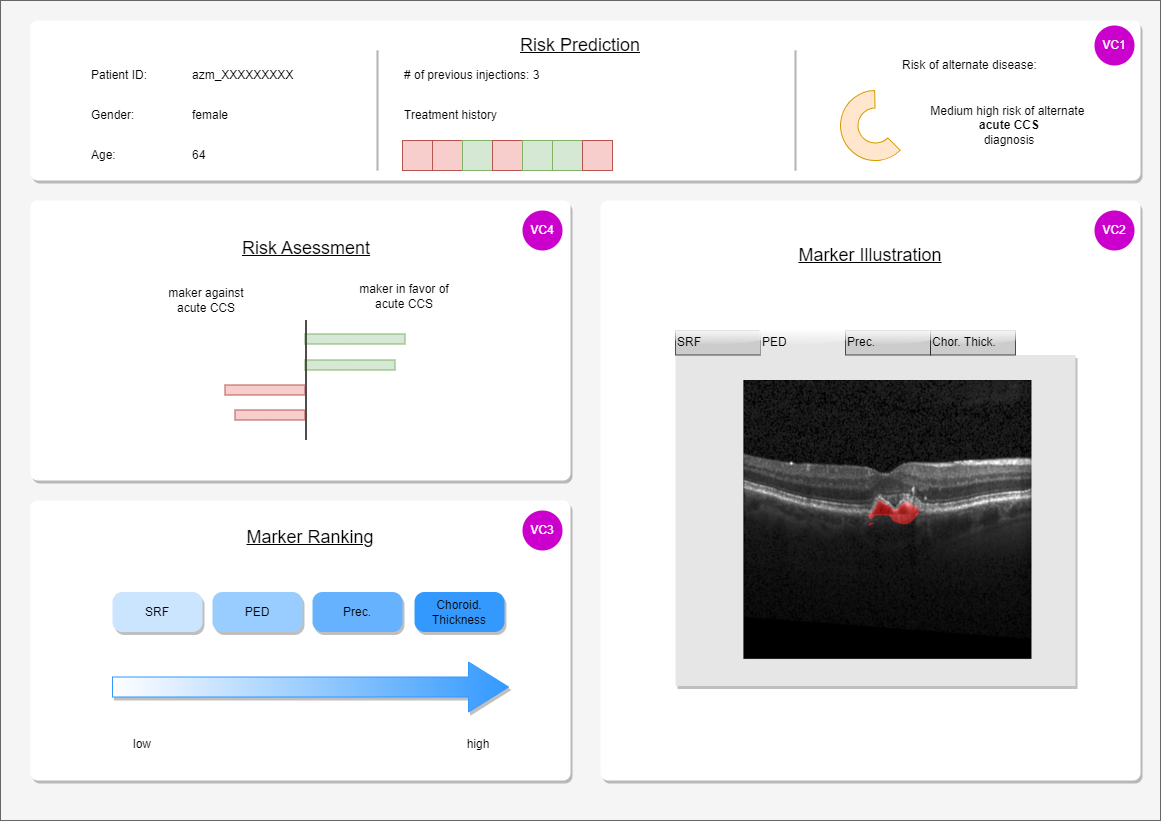}
  \caption{
            Proposed interactive DSS prototype to differentiate between AMD and other mis-diagnosed diseases, the pink circles indicate individual visual components (VC).
            VC1 provides an overall patient overview, including treatment history (which consists of injections vs no injections for AMD treatment) and an risk evaluation indicating how likely it is that the patient has been misdiagnosed.
            VC2 illustrates the individual markers with their evaluation through the system. This allows the user to investigate individual markers in-depth.
            VC4 provides a data-centric reasoning panel to illustrate the contribution of the individual markers to the risk prediction.
            VC3 is the interactive component of the demo. Here, the user may re-order the marker ranking which leads to a retraining of the risk assessment system in the back-end.
          }
  \label{img:DSS-AMD-Prototype}
\end{figure}

A study by Brinkman et al \cite{brinkmann_initiale_2019} shows that there is a significant risk of related diseases being accidentally diagnosed as AMD. 
In order to provide diagnostic support for AMD to healthcare professionals, one of the project goals is to provide a diagnostic support system (DSS).
The project's the clinical partners identified five diseases that are at risk of being diagnosed as AMD.
To support prototype development for each disease 20 images have been provided by the partners in the data warehouse.
Up to date a preliminary prototype has been developed. 
It evaluates medical markers that indicate an alternate diagnosis with the help of the segmentation of the retinal layers, provided by a DL model, and classical computer vision methods, that operate on top of the segmentation.
F.ex. to compute the choroidal thickness, the segmentation of the choroidal layer at the center of the image can be used.
In order to improve the prototype, it is desirable to incorporate more interactivity. 

In \cite{bhattacharya_directive_2023} Bhattacharya et. al explore different approaches to a data-centric design in medical DSS. 
Here, the authors developed a dashboard for monitoring diabetes onset, aimed at healthcare professionals and patients.
While the main focus of this work lies on exploring the impact of design choices onto the usefulness of the application, one notable feature are the what-if explorations.
These allow the users to explore the effects of the input parameters (such as waist circumference, alcohol intake, etc.) onto the predicted diabetes onset.

Our prototype providing risk assessment for misdiagnosing AMD is aimed at healthcare professionals only.
It is founded on marker-based reasoning, thus providing a reasoning-framework that healthcare professionals are accustomed to.
For the new prototype version it is planned to incorporate a marker ranking, that can be re-ordered (Figure \ref{img:DSS-AMD-Prototype} visual component number 3).
This allows the user to better understand the reasoning behind the risk prediction, which can be adjusted through drag-and-drop by the user, incorporating their feedback and promoting the interactive aspect of the tool.

\section{Discussion}

Our AL framework MedDeepCyleAL provides a framework for AL experiments that supports segmentation and classification in medical images. 
It can easily be used with different deep learning architectures and self-supervised models can be incorporated. 
Different experiments are planned with regard to AL.

\subsection{Partial Labeling}

Here, we hypothesize that some parts of the retinal images of AMD patients are easier to segment than others. 
While the upper-most layer is easily distinguished from the background/vitreous eye, distinguishing neighboring layers is more challenging.
We are training a DL model on retinal OCT images, whose input is a 2D image of the retina and that creates a (semantic) segmentation of the input image. 
While the common approach involves training a model on a fully labeled image, our approach only requires labeling in specific areas. 
Which areas will be presented to the annotator for labeling is determined through AL.
Note that in most AL approaches, the AL algorithm would be used to compile a training set for a deep learning model. 
In this experiment, we use a pre-defined, fixed set of images.
With the help of AL, it is determined what layers/medical markers of an individual image will be annotated. 
Instead of training the model with a fully annotated ground truth, we use a weighted loss that effectively masks out non-annotated areas.
Thus, we reduce annotation cost not by reducing the required data set size to attain a certain score but reduces the amount of annotations that need to be added to individual images.

\subsection{Active Selection with Self-Supervised Learning}

Along with active learning, another paradigm to reduce the annotation effort is self-supervised training that learns from a large pool of unlabeled data in an unsupervised way. Self-supervised learning (SSL) has gained attention as it can utilize large amounts of unlabeled data, which are often readily available. It leverages the inherent structure or information present in the input data itself to create supervised learning tasks. The model learns to predict certain aspects of the data, such as filling in missing parts, generating transformed versions, or clustering group of similar samples closer and diverse samples afar (contrastrive learning) \cite{balestriero2023cookbook}. Recent studies suggest using self-supervised learning with active learning can be a powerful combination that leverages the benefits of both approaches \cite{bengar2021reducing}. Neural models for our task can be pre-trained using large pool of unlabeled OCT slices by applying self-supervised tasks in order to learn meaningful representations of the data. And by integrating SSL optimization in the sample selection of our active learning framework, we plan to evaluate their performance on medical imaging domain.

\subsection{The Benefit of Active Learning for the Annotation Process}

In AL, the models that are gradually trained, are sometimes used to provide suggestions to the annotators in order to speed up the annotation process. 
In this experiment, we measure the time-wise benefit of such an approach for the segmentation of retinal layers. 
The underlying hypothesis is that, while the models trained on the partial data set may not be as accurate, suggestions generated with their help still provide a speed-up in the annotation process.  
Since the purpose of uncertainty sampling in AL is to provide challenging examples, this benefit has to be present for challenging data points. 

We expect this effect to take place the moment the provided suggestions are exact enough, such that they can be corrected with an effort that falls below the creation of the same annotation from the very beginning, or that the suggestions are good enough for some annotation classes, such that the annotator may skip the particular example at all.
A known challenge from previous studies is the distinction between secondary effects and the actual benefit of the described method. 
For one it is difficult to measure the inter-annotator difference and then it provided to be difficult to differentiate between the learning effect of performing the annotation process (f.ex. learning to use the tool) and benefit of the described method.
We are countering the former by choosing an appropriate amount of annotators and the latter by giving the annotators some example annotations to get to know the tool.

\section*{Acknowledgement}
The funding for this project was provided in part by the German Federal Ministry of Education and Research (BMBF) under grant number 16SV8639 (Ophthalmo-AI). Additional support was received from the Lower Saxony Ministry of Science and Culture and the Endowed Chair of Applied Artificial Intelligence (AAI) at the University of Oldenburg.

\newpage
\appendix

\section{Annotation Tool Protocol for Annotating the Inner Limiting Membrane Layer}
\label{annotationtool}

\begin{center}
\captionof{lstlisting}{Example protocol consisting of a load state loading an OCT\, an image segmentation task for the Inner Limiting Membrane, a categorial question for the type of Macular Foramen (none, pseudo lammelar or full-thickness), and a summary state\, providing a summary of all annotations at the end.}
\begin{lstlisting}[language=yaml, numbers=none]
start:
    type: load
    dataloader: OCTLoader
    transitions:
        - next:
              target: seg_ilm
seg_ilm:
    type: octSegmentation
    question: Inner Limiting Membrane
    annotation_type: line
    transitions:
        - "*":
              target: macular_foramen
macular_foramen:
    question: Macular Foramen
    type: select
    options:
        - '-'
        - pseudo
        - lamellar
        - full-thickness
    transitions:
        - '*':
              target: summary
summary:
    type: summary_oct
    question: Summary
    transitions:
        - '*':
              target: end
\end{lstlisting}
\end{center}
\section{Example Annotation Workflow}
\label{workflow}

\begin{figure}[H]
    \centering
    \includegraphics[width=0.9\textwidth]{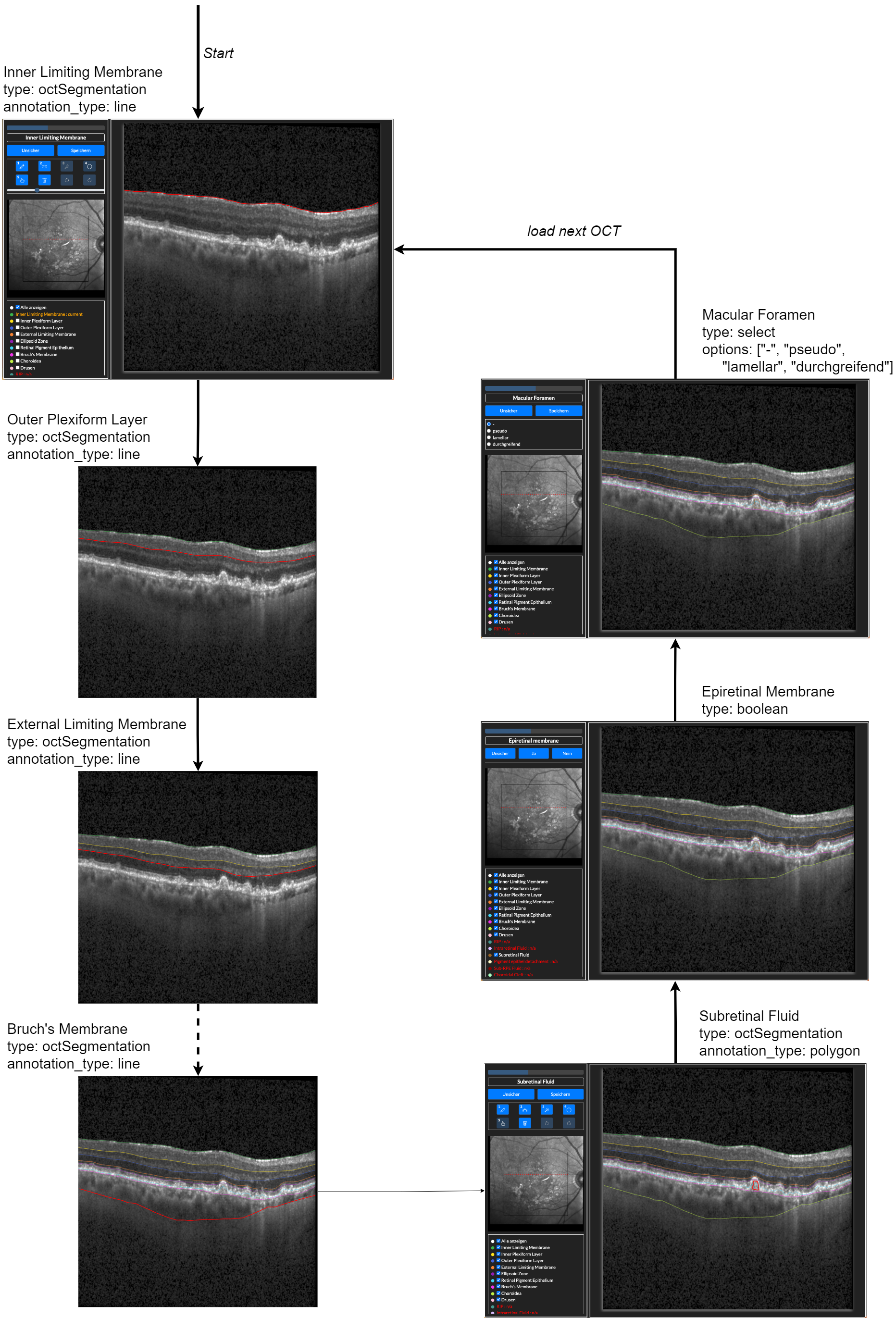}
\end{figure}

\newpage
\bibliographystyle{IEEEtranN}
\bibliography{references}

\end{document}